\documentclass{article}
\usepackage{iclr2027_conference,times}
\iclrfinalcopy

\usepackage{amsmath,amsfonts,bm}

\def\eqref#1{(\ref{#1})}

\def\1{\bm{1}}

\DeclareMathAlphabet{\mathsfit}{\encodingdefault}{\sfdefault}{m}{sl}
\SetMathAlphabet{\mathsfit}{bold}{\encodingdefault}{\sfdefault}{bx}{n}

\usepackage{url}
\usepackage{algorithm}
\usepackage{algorithmic}
\usepackage{booktabs}
\usepackage{graphicx}
\usepackage{xcolor}
\usepackage{colortbl}
\usepackage{multirow}
\usepackage{hyperref}

\title{GLF-Q: Global-Local Feature-based Quantization for Vision Transformers}

\author{%
Peilin Sun\textsuperscript{1,2} \quad Guang Liang\textsuperscript{1,2,3} \quad Jin Tong\textsuperscript{1,2} \quad Jianxin Wu\textsuperscript{1,2}\thanks{Corresponding author.}\\
{\normalfont\small \textsuperscript{1}State Key Laboratory of Novel Software Technology, Nanjing University, Nanjing 210023, China}\\
{\normalfont\small \textsuperscript{2}School of Artificial Intelligence, Nanjing University, Nanjing 210023, China}\\
{\normalfont\small \textsuperscript{3}Zhongguancun Academy, Beijing 100094, China}\\
{\normalfont\small\texttt{\{sunpl, liangg, tongj\}@lamda.nju.edu.cn, wujx2001@nju.edu.cn}}
}

\begin{document}
\maketitle
\lhead{\footnotesize GLF-Q: Global-Local Feature-based Quantization for Vision Transformers}

\begin{abstract}
Post-training quantization (PTQ) efficiently compresses Vision Transformers (ViTs) without retraining, yet suffers severe accuracy degradation at low bit-widths. Existing optimization-based PTQ methods guide block reconstruction via either soft logits or second-order Hessian proxies. Logit supervision is prone to overfitting on limited calibration data, while Hessian approximations incur structural truncation errors. To address these limitations, we propose \textbf{GLF-Q}, a novel PTQ framework guided by Global-Local Feature alignment. GLF-Q propagates quantized block outputs through downstream full-precision layers to align penultimate-layer representations under local output regularization, providing downstream feature supervision without explicitly approximating the Hessian or using a Taylor expansion. Furthermore, offline Hadamard transformations are introduced with zero runtime overhead to disperse activation outliers across channels, effectively contracting dynamic ranges and reducing quantization errors. Meanwhile, optimizing this loss via a Straight-Through Estimator (STE) achieves rapid convergence, bypassing continuous relaxation rounding formulations such as AdaRound. Extensive experiments across representative ViT architectures demonstrate that GLF-Q with standard uniform quantizers substantially outperforms state-of-the-art methods under 3-bit quantization on image classification. In addition, GLF-Q exhibits strong out-of-domain calibration robustness and achieves speedups under 8-bit GPU deployment.
\end{abstract}

\section{Introduction}
\label{sec:intro}

Vision Transformers (ViTs)~\citep{dosovitskiy2020image} have achieved remarkable success across diverse visual recognition tasks, such as image classification~\citep{touvron2021training}, as well as object detection and instance segmentation~\citep{he2017mask,cai2018cascade,liu2021swin}, owing to their powerful capability in modeling long-range dependencies. However, their massive model sizes, intensive self-attention computations, and heavy memory bandwidth demands severely limit practical deployment on resource-constrained edge devices. Among various model compression techniques, network quantization has attracted widespread attention, as it converts floating-point weights and activations into low-bit representations, thereby significantly reducing memory footprints and substantially accelerating hardware inference simultaneously.

Existing quantization approaches can be broadly divided into Quantization-Aware Training (QAT) and Post-Training Quantization (PTQ). Although QAT can effectively recover model accuracy through retraining, it requires full labeled data and expensive retraining, severely restricting its practical deployment. In contrast, PTQ requires only a small amount of unlabeled data for calibration, making it a practical and efficient approach.

Depending on whether optimization is required, existing PTQ methods primarily fall into two categories: calibration-only methods~\citep{yuan2022ptq4vit,li2023repq,ding2022towards} and optimization-based methods~\citep{nagel2020adaround,li2021brecq,liu2023pd,wu2025fima,wu2025aphq,hwang2026ls}. Calibration-only methods adjust quantization parameters solely during the calibration phase. However, in low-bit settings, quantization errors increase substantially, leading to severe accuracy degradation. To overcome this limitation, optimization-based methods perform block-wise reconstruction to compensate for quantization loss, emerging as the prevailing paradigm for ViT PTQ.

In block-wise reconstruction, the choice of supervision signal directly affects the accuracy of the quantized model. Existing reconstruction approaches can be broadly classified into two categories based on their supervisory signals. The first category comprises logit-guided methods~\citep{liu2023pd}, which seek global supervision by minimizing the prediction difference on output soft logits at the task head. However, fitting output logits on a small calibration set can lead to overfitting and limit generalization, as shown in Appendix~\ref{sec:appendix_overfitting}.

Hessian-guided reconstruction~\citep{li2021brecq,wu2025fima,wu2025aphq,hwang2026ls} uses second-order loss approximations to weight reconstruction errors according to their estimated impact on the task loss. Since computing the full Hessian is expensive, these methods employ tractable Hessian approximations, such as diagonal or low-rank representations. The accuracy of the estimated loss change depends on both the Hessian approximation and the second-order Taylor approximation. Under low-bit quantization, larger perturbations may make higher-order effects more relevant and reduce the accuracy of the second-order approximation.

These limitations motivate evaluating feature differences directly through the full-precision downstream network, without explicitly constructing a Hessian approximation or truncating a Taylor expansion. We combine penultimate-layer feature alignment with local reconstruction regularization to improve generalization on limited calibration data.

Motivated by these insights, we propose GLF-Q, an efficient post-training quantization framework for Vision Transformers based on Global-Local Feature alignment. GLF-Q introduces a \emph{Global-Local Feature (GLF) loss}, which combines penultimate feature alignment with local reconstruction regularization. Furthermore, to address the critical challenge of extreme activation outliers and inter-channel scale disparities in ViTs, we systematically introduce \emph{offline Hadamard transformations}~\citep{ashkboos2024quarot} into the ViT PTQ pipeline for the first time. This cheap operation evenly distributes energy concentrated in outlier channels across all dimensions, contracting activation dynamic ranges to effectively suppress quantization errors with zero runtime overhead. In addition, regarding the reconstruction optimization mechanism, GLF-Q avoids continuous rounding relaxation methods such as AdaRound~\citep{nagel2020adaround}. Instead, it uses a basic Straight-Through Estimator (STE)~\citep{bengio2013estimating} to achieve fast and stable convergence, thereby improving reconstruction efficiency.

Our main contributions are summarized as follows:
\begin{itemize}
    \item We propose the Global-Local Feature (GLF) loss, which combines global feature alignment with local reconstruction regularization to mitigate calibration overfitting and avoid explicit Hessian approximations.
    \item To mitigate activation outliers, we introduce offline Hadamard transformations to spread large activation values across channels, reducing quantization errors. For reconstruction optimization, we use a simple STE instead of continuous rounding relaxation to achieve fast convergence.
    \item We demonstrate practical hardware efficiency and strong out-of-domain generalizability. GLF-Q achieves higher accuracy and lower inference latency under practical 8-bit GPU deployment. It also maintains strong accuracy under out-of-domain calibration.
\end{itemize}

\section{Related Work}
\label{sec:related}

Model quantization~\citep{gholami2022survey} converts floating-point weights and activations into low-bit representations to reduce memory usage and accelerate inference. Existing quantization methods can be classified into Quantization-Aware Training (QAT), which requires retraining, and Post-Training Quantization (PTQ), which does not. Although QAT methods~\citep{esser2019learned,li2022q,liang2026gplq} achieve competitive accuracy, their reliance on full labeled data and expensive retraining severely limit practical deployment.

In contrast, PTQ calibrates models using only a small set of unlabeled data without retraining, making it an efficient and popular approach. Depending on whether optimization is involved, PTQ methods broadly fall into calibration-only methods~\citep{yuan2022ptq4vit,ding2022towards,li2023repq,wu2024adalog,zhong2024erq,moon2024instance,fu2025quantization,jiang2026uq} and optimization-based methods~\citep{nagel2020adaround,li2021brecq,wei2022qdrop,liu2023pd,zhong2023s,yang2024dopq,ma2024outlier,wu2025aphq,wu2025fima,hwang2026ls}.

Calibration-only methods determine quantization parameters directly during a forward calibration pass without backpropagation. RepQ-ViT~\citep{li2023repq} uses scale reparameterization to convert complex activation quantizers into hardware-friendly forms for efficient inference. AdaLog~\citep{wu2024adalog} addresses heavy-tailed activations via dynamic log-base tuning paired with a progressive search strategy. UQ-ViT~\citep{jiang2026uq} introduces DeMax and NormQuant to accommodate extreme activation distributions with uniform quantizers. Despite these efforts, calibration-only methods suffer substantial accuracy loss in low-bit settings, while those employing specialized non-uniform formats remain difficult to deploy on general hardware.

To recover accuracy, optimization-based methods perform block-wise reconstruction to reduce quantization loss. BRECQ~\citep{li2021brecq} optimizes weight rounding using second-order error analysis across residual blocks. QDrop~\citep{wei2022qdrop} randomly drops activation quantization during reconstruction to enhance model flatness and noise tolerance. PD-Quant~\citep{liu2023pd} minimizes prediction differences on soft classification logits for global guidance. DopQ-ViT~\citep{yang2024dopq} handles activation outliers during reconstruction via LayerNorm input reparameterization and specialized non-uniform quantizers. APHQ-ViT~\citep{wu2025aphq} couples an average perturbation Hessian loss with MLP reconstruction. FIMA-Q~\citep{wu2025fima} constructs a composite Hessian loss by modeling both diagonal and low-rank cross-channel dependencies. LS-ViT~\citep{hwang2026ls} formulates Hessian proxy estimation as a least-squares problem over calibration gradients, achieving rapid reconstruction with single-pass backpropagation. Despite these advances, existing reconstruction methods still face limitations. Logit-guided methods can overfit limited calibration data, while Hessian-guided methods rely on approximations that may become less accurate under low-bit quantization.

\section{Method}
\label{sec:method}

\begin{figure}[t]
\centering
\includegraphics[width=\linewidth]{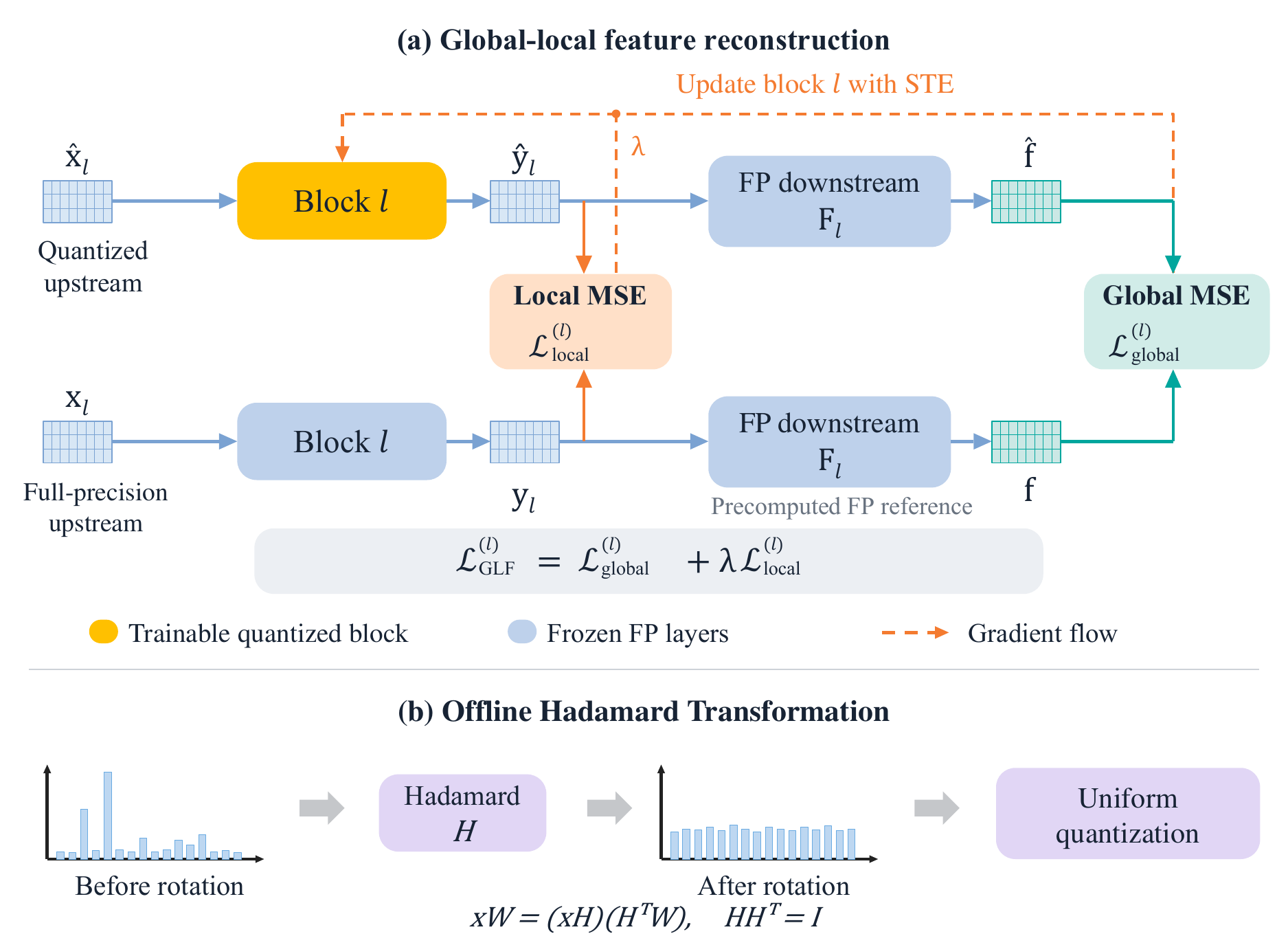}
\caption{Overview of the proposed GLF-Q framework. (a) Global-local feature reconstruction. For block $l$, the quantized block output $\hat{y}_l$ is propagated through the frozen downstream full-precision network $F_l$ to match the precomputed full-precision reference feature $f$, under local reconstruction regularization. Gradients are backpropagated directly to update the quantized block via STE. (b) Offline Hadamard transformation. Orthogonal transformations are fused offline into adjacent linear weights to suppress activation outliers and balance channel distributions with zero runtime overhead.}
\label{fig:overview}
\end{figure}

\subsection{Preliminaries and Motivation}
\label{sec:preliminaries}

\paragraph{Logit-Guided Block Reconstruction.}
In block-wise reconstruction, parameter optimization relies heavily on the supervisory signal. To incorporate global task guidance, logit-guided methods, represented by PD-Quant~\citep{liu2023pd}, minimize the prediction difference on output soft logits at the final task head. However, fitting output logits on a small calibration set can lead to overfitting and limit generalization.

\paragraph{Hessian-Guided Block Reconstruction.}
Let $\theta$ denote the parameters of the pretrained model and $\Delta\theta$ be the quantization perturbation. The change in task loss can be approximated via a second-order Taylor expansion:
\begin{equation}
\mathbb{E}\left[\mathcal{L}(\theta+\Delta\theta)\right] - \mathbb{E}\left[\mathcal{L}(\theta)\right] \approx \mathbb{E}\left[\Delta\theta^{\top} g^{(\theta)} + \frac{1}{2}\Delta\theta^{\top} H^{(\theta)}\Delta\theta\right],
\end{equation}
where $g^{(\theta)} = \nabla_{\theta}\mathcal{L}$ is the gradient and $H^{(\theta)} = \nabla_{\theta}^{2}\mathcal{L}$ is the Hessian matrix. Since a well-converged model satisfies $g^{(\theta)} \approx 0$, existing reconstruction-based PTQ methods map parameter perturbations to the module output perturbation $\Delta y$ via the chain rule:
\begin{equation}
\mathcal{L}_{\mathrm{H}}(\Delta y) = \frac{1}{2}\Delta y^{\top} H^{(y)}\Delta y,
\end{equation}
where $H^{(y)} = \nabla_{y}^{2}\mathcal{L}$ is the Hessian matrix with respect to the module output $y$. Compared with vanilla MSE, $\mathcal{L}_{\mathrm{H}}$ weights quantization errors according to directional task sensitivity.

Because computing the dense matrix $H^{(y)}$ is intractable, existing methods commonly adopt diagonal, low-rank, or Fisher matrix approximations~\citep{li2021brecq,wu2025fima,wu2025aphq,hwang2026ls}. However, the accuracy of these proxies depends on the second-order Taylor approximation and the structural assumptions used to estimate curvature. Under low-bit quantization, larger perturbations may reduce the accuracy of the local quadratic approximation, motivating direct measurement of downstream feature differences.

\subsection{Global-Local Feature Loss}
\label{sec:gfm}

To address the two limitations discussed above, we introduce the Global-Local Feature (GLF) loss, illustrated in Figure~\ref{fig:overview}(a). Specifically, the GLF loss directly leverages global penultimate-layer feature representations~\citep{wang2021distilling} to evaluate downstream quantization loss under local regularization.

When reconstructing block $l$, preceding quantized blocks are frozen, with their cached activations serving as input $\hat{x}_l$. We cache the global feature $f \in \mathbb{R}^C$ and local block output $y_l \in \mathbb{R}^{N \times C}$ on $\mathcal{D}_{\mathrm{cal}}$ using the full-precision teacher after offline transformations. During reconstruction, the dequantized block output $\hat{y}_l \in \mathbb{R}^{N \times C}$ is dynamically computed from $\hat{x}_l$ and fed into the frozen downstream full-precision suffix $F_l$:
\begin{equation}
\hat{f} = F_l(\hat{y}_l),
\end{equation}
where $\hat{f} \in \mathbb{R}^C$ denotes the global penultimate representation immediately preceding the task head. For image classification, $f$ is the final class token representation for ViT and DeiT, and the globally averaged token representation for Swin. The global feature loss is formulated as
\begin{equation}
\label{eq:global_feature_loss}
\mathcal{L}_{\mathrm{global}}^{(l)} = \mathbb{E}_{x \sim \mathcal{D}_{\mathrm{cal}}} \left[ \frac{1}{C} \left\| \hat{f} - f \right\|_2^2 \right].
\end{equation}

This loss measures the feature differences caused by quantization after propagation through the full-precision downstream network. Concurrently, we introduce the local feature loss as a regularizer:
\begin{equation}
\label{eq:local_feature_loss}
\mathcal{L}_{\mathrm{local}}^{(l)} = \mathbb{E}_{x \sim \mathcal{D}_{\mathrm{cal}}} \left[ \frac{1}{NC} \left\| \hat{y}_l - y_l \right\|_F^2 \right].
\end{equation}
The GLF loss for block $l$ is defined as
\begin{equation}
\label{eq:glf_loss}
\mathcal{L}_{\mathrm{GLF}}^{(l)} = \mathcal{L}_{\mathrm{global}}^{(l)} + \lambda \mathcal{L}_{\mathrm{local}}^{(l)},
\end{equation}
where $\lambda > 0$ is the regularization coefficient controlling the strength of local regularization. In practice, each loss term is normalized by its value on the first reconstruction minibatch, with the normalization factors fixed throughout optimization.

During optimization of block $l$, the downstream suffix $F_l$ remains frozen, yet gradients backpropagate through $F_l$ via STE to update this block's parameters $\theta_l$ and quantization scale $s_l$. Once reconstructed, block $l$ is frozen in its quantized state, and the pipeline advances sequentially to the next block.

\subsection{GLF-Q Framework}
\label{sec:pipeline}

As shown in Figure~\ref{fig:overview} and Algorithm~\ref{alg:glfq_pipeline} in Appendix~\ref{sec:appendix_pipeline}, GLF-Q follows a block-wise quantization pipeline. First, we apply offline Hadamard transformations to mitigate activation outliers with zero runtime overhead. We then perform MLP reconstruction by replacing GELU with strictly non-negative ReLU to expand the effective dynamic range. Finally, we carry out block reconstruction guided by the GLF loss.

\paragraph{Offline Hadamard Transformation.}
Directly quantizing Vision Transformers remains challenged by channel scale disparities and activation outliers~\citep{darcet2023vision,sun2026nonlinear}, which widen the dynamic range and increase quantization errors. To disperse activation outliers across channels without additional inference overhead, we apply an offline Hadamard transformation~\citep{ashkboos2024quarot} prior to reconstruction, as depicted in Figure~\ref{fig:overview}(b). Specifically, let $M = I - \frac{1}{C}\mathbf{1}\mathbf{1}^{\top} \in \mathbb{R}^{C \times C}$ denote the mean-subtraction projection matrix, where $I$ is the identity matrix and $\mathbf{1} \in \mathbb{R}^{C}$ is the all-ones vector, such that the centered activation satisfies $x_{\mathrm{c}} = xM$. Following SliceGPT~\citep{ashkboos2024slicegpt}, both the mean centering and affine operations of LayerNorm are absorbed into adjacent linear layers:
\begin{equation}
\operatorname{LN}(x)W + b = \operatorname{RMSNorm}(x_{\mathrm{c}})\widetilde{W} + \widetilde{b},
\end{equation}
where the preceding layer absorbs mean-centering:
\begin{equation}
\widetilde{W}_{\mathrm{prev}} = W_{\mathrm{prev}}M, \qquad \widetilde{b}_{\mathrm{prev}} = b_{\mathrm{prev}}M,
\end{equation}
and the subsequent layer absorbs the scaling $\gamma$ and bias $\beta$:
\begin{equation}
\widetilde{W} = \operatorname{Diag}(\gamma)W, \qquad \widetilde{b} = \beta W + b.
\end{equation}
This transformation converts LayerNorm into standard RMSNorm. We then apply a randomized Hadamard transform $R_1$ to rotate the residual stream:
\begin{equation}
xW = (xR_1)(R_1^{\top}W),
\end{equation}
where $R_1R_1^{\top} = I$. Both $R_1$ and $R_1^{\top}$ are fused into weights offline before quantization, distributing activation outliers across channels. For multi-head self-attention, head-wise orthogonal rotations are applied to QK and VO projections to eliminate outliers without cross-head interference. Full network equivalence, coordinate alignment, and generalized Kronecker constructions for non-$2^k$ dimensions are detailed in Appendix~\ref{sec:appendix_hadamard}.

\paragraph{MLP Reconstruction.}
For nonlinear activations, the asymmetric negative tail of GELU wastes the dynamic range of low-bit quantizers and causes quantization loss. Following APHQ-ViT~\citep{wu2025aphq}, before block reconstruction, we conduct an MLP reconstruction step. We replace GELU with ReLU in the MLP module to avoid allocating quantization levels to negative values, and then reconstruct the MLP to reduce the changes caused by activation replacement.

\paragraph{Efficient Block Reconstruction via STE.}
Unlike conventional reconstruction methods that introduce continuous relaxation variables and require tens of thousands of iterations, such as AdaRound~\citep{nagel2020adaround}, we directly optimize full-precision weights $\theta_l$ and quantization scales $s_l$ using STE~\citep{bengio2013estimating} under the GLF loss in Eq.~\eqref{eq:glf_loss}. This direct formulation converges within only 3000 iterations, significantly accelerating reconstruction while achieving superior low-bit accuracy.

\begin{table*}[t]
\caption{
Top-1 accuracy comparison across representative Vision Transformers on ImageNet. 
``*'' denotes results reimplemented using the FIMA-Q framework, as QDrop and PD-Quant were originally designed for CNNs. 
``Opt.'' indicates whether a PTQ method is optimization-based ($\checkmark$) or calibration-only ($\times$). 
``Spec.'' indicates reliance on a specialized non-uniform quantizer ($\checkmark$) versus a standard uniform quantizer ($\times$). 
The best results within each bit-width setting are highlighted in \textbf{boldface}.
}
\label{tab:ImageNet}
\begin{center}
\resizebox{\textwidth}{!}{%
\begin{tabular}{ccccccccccc}
    \toprule
    \textbf{Method} & \textbf{Opt.} & \textbf{Spec.} & \textbf{W/A} & \textbf{ViT-S} & \textbf{ViT-B} & \textbf{DeiT-T} & \textbf{DeiT-S} & \textbf{DeiT-B} & \textbf{Swin-S} & \textbf{Swin-B} \\
    \midrule
    Full-Prec & - & - & 32/32 & 81.39 & 84.54 & 72.21 & 79.85 & 81.80 & 83.23 & 85.27 \\
    \midrule
    PTQ4ViT~\citep{yuan2022ptq4vit} & $\times$ & $\checkmark$ & 3/3 & 0.10 & 0.10 & 3.50 & 0.10 & 31.06 & 28.69 & 20.13 \\
    RepQ-ViT~\citep{li2023repq} & $\times$ & $\checkmark$ & 3/3 & 0.10 & 0.10 & 0.10 & 0.10 & 0.10 & 0.10 & 0.10 \\
    AdaLog~\citep{wu2024adalog} & $\times$ & $\checkmark$ & 3/3 & 13.88 & 37.91 & 31.56 & 24.47 & 57.47 & 64.41 & 69.75 \\
    I\&S-ViT~\citep{zhong2023s} & $\checkmark$ & $\checkmark$ & 3/3 & 45.16 & 63.77 & 41.52 & 55.78 & 73.30 & 74.20 & 69.30 \\
    DopQ-ViT~\citep{yang2024dopq} & $\checkmark$ & $\checkmark$ & 3/3 & 54.72 & 65.76 & 44.71 & 59.26 & 74.91 & 74.77 & 69.63 \\
    QDrop*~\citep{wei2022qdrop} & $\checkmark$ & $\times$ & 3/3 & 41.05 & 74.75 & 46.88 & 50.95 &  72.97&  74.67&  76.57\\
    PD-Quant*~\citep{liu2023pd} & $\checkmark$ & $\times$ & 3/3 & 40.52 & 75.72 & 53.23 & 60.72 & 74.54 & 74.59 & 76.71 \\
    APHQ-ViT~\citep{wu2025aphq} & $\checkmark$ & $\times$ & 3/3 & 63.17 & 76.31 & 55.42 & 68.76 & 76.31 & 76.10 & 78.14 \\
    FIMA-Q~\citep{wu2025fima}& $\checkmark$ & $\times$ & 3/3 & 64.09&  77.63& 55.55& 69.13&  76.54&  77.26&  78.82\\
    LS-ViT~\citep{hwang2026ls} & $\checkmark$ & $\times$ & 3/3 & 64.10 & 77.65 & 55.72 & 69.41 & 76.57 & 77.39 & 79.40 \\
    \textbf{GLF-Q (Ours)} & $\checkmark$ & $\times$ & 3/3 & \textbf{67.40} & \textbf{78.18} & \textbf{56.48} & \textbf{71.55} & \textbf{77.20} & \textbf{78.10} & \textbf{80.45} \\
    \midrule
    PTQ4ViT~\citep{yuan2022ptq4vit} & $\times$ & $\checkmark$ & 4/4 & 42.57 & 30.69 & 36.96 & 34.08 & 64.39 & 76.09 & 74.02  \\
    APQ-ViT~\citep{ding2022towards} & $\times$ & $\checkmark$ & 4/4 & 47.95 & 41.41 & 47.94 & 43.55 & 67.48 & 77.15 & 76.48 \\
    RepQ-ViT~\citep{li2023repq}& $\times$ & $\checkmark$ & 4/4 & 65.05 & 68.48 & 57.43 & 69.03 & 75.61 & 79.45 & 78.32 \\
    ERQ~\citep{zhong2024erq} & $\times$ & $\checkmark$ & 4/4 &68.91 & 76.63 & 60.29 & 72.56 & 78.23 & 80.74 & 82.44 \\
    IGQ-ViT~\citep{moon2024instance} & $\times$ & $\checkmark$ & 4/4 & 73.61 & 79.32 & 62.45 & 74.66 & 79.23 & 80.98 & 83.14 \\
    AdaLog~\citep{wu2024adalog} & $\times$ & $\checkmark$ & 4/4 & 72.75 & 79.68 & 63.52 & 72.06 & 78.03 & 80.77 & 82.47 \\
    UQ-ViT~\citep{jiang2026uq} & $\times$ & $\times$ & 4/4 & 68.34 & 72.07 & 59.71 & 72.13 & 76.59 & 79.93 & 81.96 \\
    I\&S-ViT ~\citep{zhong2023s} & $\checkmark$ & $\checkmark$ & 4/4 & 74.87 & 80.07 & 65.21 & 75.81 & 79.97 & 81.17 & 82.60 \\
    DopQ-ViT~\citep{yang2024dopq} & $\checkmark$ & $\checkmark$ & 4/4 & 75.69 & 80.95 & 65.54 & 75.84 & 80.13 & 81.71 & 83.34 \\
    QDrop*~\citep{wei2022qdrop} & $\checkmark$ & $\times$ & 4/4 & 71.84 &  82.63& 65.27 &  72.64&  79.96&  81.21&  82.99\\
    PD-Quant*~\citep{liu2023pd} & $\checkmark$ & $\times$ & 4/4 & 72.98 & 82.72 & 66.23 & 74.98 & 79.90 & 81.28 & 82.92 \\
    OASQ~\citep{ma2024outlier} & \checkmark & \(\times\) & 4/4 & 72.88 & 76.59 & 66.31 & 76.00 & 78.83 & 81.02 & 82.46 \\
    APHQ-ViT~\citep{wu2025aphq} & $\checkmark$ & $\times$ & 4/4 & 76.07 & 82.41 & 66.66 & 76.40 & 80.21 & 81.81 & 83.42 \\
    FIMA-Q~\citep{wu2025fima}& $\checkmark$ & $\times$ & 4/4 & 76.68&  83.04& 66.84&  76.87&  80.33&  81.82&  83.60\\
    LS-ViT~\citep{hwang2026ls} & $\checkmark$ & $\times$ & 4/4 & 76.67 & 83.08 & 67.05 & 76.89 & 80.41 & 81.82 & 83.62 \\
    \textbf{GLF-Q (Ours)} & $\checkmark$ & $\times$ & 4/4 & \textbf{77.82} & \textbf{83.22} & \textbf{67.37} & \textbf{77.34} & \textbf{80.53} & \textbf{82.01} & \textbf{83.80} \\
   \bottomrule
\end{tabular}%
}
\end{center}
\end{table*}

\section{Experiments}
\label{sec:experiments}

We evaluate GLF-Q on image classification, object detection, and instance segmentation. We further conduct ablation studies, assess out-of-domain calibration robustness, and compare inference latency and training time. Additional experiments are provided in the appendix.

\subsection{Experimental Setup}
\label{sec:exp_setup}

\paragraph{Datasets and Models.}
For image classification, we evaluate on ImageNet~\citep{deng2009imagenet} across ViT~\citep{dosovitskiy2020image}, DeiT~\citep{touvron2021training}, and Swin~\citep{liu2021swin}. For object detection and instance segmentation, we conduct experiments on COCO~\citep{lin2014microsoft} using Mask R-CNN~\citep{he2017mask} and Cascade Mask R-CNN~\citep{cai2018cascade} with Swin backbones. All pretrained full-precision Vision Transformers are obtained from the \texttt{timm} library\footnote{\url{https://github.com/huggingface/pytorch-image-models}}, while pretrained detection and segmentation models are sourced from MMDetection~\citep{chen2019mmdetection}.

\paragraph{Implementation Details.}
We apply standard channel-wise uniform quantization to weights and layer-wise uniform quantization to activations, including Softmax outputs, as detailed in Appendix~\ref{sec:appendix_uniform_quant}. Following reconstruction-based PTQ methods~\citep{li2021brecq,wei2022qdrop,wu2025aphq,wu2025fima}, we randomly sample 1024 unlabeled images from ImageNet for image classification and 256 unlabeled images from COCO for object detection and instance segmentation as the calibration sets. We set the batch size, learning rate for activation quantization, learning rate for tuning weight, and reconstruction iterations as 32, 2e-4, 2e-5, and 3000, respectively. The regularization coefficient $\lambda$ in Eq.~\eqref{eq:glf_loss} is set to $2$ across all experiments. For detection and segmentation, the global term averages the normalized bbox and mask RoI feature losses. For Swin, where hierarchical patch merging hinders exact RMSNorm conversion, we retain LayerNorm without residual Hadamard, pairing head-wise QK/VO Hadamard rotations with LayerNorm input reparameterization~\citep{yang2024dopq}. More implementation details are provided in Appendix~\ref{sec:appendix_implementation}.

\subsection{Experiments on Image Classification}
\label{sec:exp_cls}

We first evaluate the performance of our method on the classification task on ImageNet in terms of Top-1 accuracy, compared to the state-of-the-art PTQ approaches. We report results across various representative Transformer architectures, including ViT, DeiT, and Swin, under 4-bit and 3-bit settings. 

As displayed in Table~\ref{tab:ImageNet}, GLF-Q achieves strong performance across diverse architectures, with particularly significant gains under 3-bit quantization. Concretely, for 4-bit quantization, while some existing methods suffer noticeable accuracy degradation, our method maintains robust performance across all architectures. Under the challenging 3-bit quantization, the performance of competing methods degrades severely, with methods such as PTQ4ViT~\citep{yuan2022ptq4vit} and RepQ-ViT~\citep{li2023repq} suffering severe accuracy degradation. In comparison, our proposed GLF-Q maintains robust accuracy, outperforming the second-best approach by 3.30\% on ViT-S.

It is worth noting that many compared approaches such as PTQ4ViT~\citep{yuan2022ptq4vit}, RepQ-ViT~\citep{li2023repq}, AdaLog~\citep{wu2024adalog}, and DopQ-ViT~\citep{yang2024dopq} attempt to boost performance by designing specific quantizers, which however are generally difficult to implement on hardware in practice. In contrast, our method relies strictly on a standard uniform quantizer while achieving superior accuracy, facilitating hardware-friendly deployment on general-purpose platforms.

\subsection{Experiments on Object Detection and Instance Segmentation}
\label{sec:exp_det}

We evaluate 4-bit quantization on COCO~\citep{lin2014microsoft} using Mask R-CNN~\citep{he2017mask} and Cascade Mask R-CNN~\citep{cai2018cascade} with Swin backbones. As shown in Table~\ref{tab:object_detection}, GLF-Q achieves competitive performance across object detection and instance segmentation, obtaining the best results on several model configurations while using standard uniform quantization. These results demonstrate the effectiveness of GLF-Q on dense prediction tasks.
\begin{table*}[t]
\caption{Object detection and instance segmentation performance on COCO with Mask R-CNN and Cascade Mask R-CNN under 4-bit quantization.}
\label{tab:object_detection}
\begin{center}
\resizebox{\textwidth}{!}{%
\begin{tabular}{lccccccccccc}
    \toprule
    \multirow{3}{*}[-0.5ex]{Method} & \multirow{3}{*}[-0.5ex]{Opt.} & \multirow{3}{*}[-0.5ex]{Spec.} & \multirow{3}{*}[-0.5ex]{W/A} & \multicolumn{4}{c}{Mask R-CNN} & \multicolumn{4}{c}{Cascade Mask R-CNN} \\
    \cmidrule(lr){5-8} \cmidrule(lr){9-12}
    & & & & \multicolumn{2}{c}{Swin-T} & \multicolumn{2}{c}{Swin-S} & \multicolumn{2}{c}{Swin-T} & \multicolumn{2}{c}{Swin-S} \\
    \cmidrule(lr){5-6} \cmidrule(lr){7-8} \cmidrule(lr){9-10} \cmidrule(lr){11-12}
    & & & & $\text{AP}^b$ & $\text{AP}^m$ & $\text{AP}^b$ & $\text{AP}^m$ & $\text{AP}^b$ & $\text{AP}^m$ & $\text{AP}^b$ & $\text{AP}^m$ \\
    \midrule
    Full-Precision & - & - & 32/32 & 46.0 & 41.6 & 48.5 & 43.3 & 50.4 & 43.7 & 51.9 & 45.0 \\
    \midrule
    PTQ4ViT~\citep{yuan2022ptq4vit} & $\times$ & $\checkmark$ & 4/4 & 6.9 & 7.0 & 26.7 & 26.6 & 14.7 & 13.5 & 0.5 & 0.5 \\
    APQ-ViT~\citep{ding2022towards} & $\times$ & $\checkmark$ & 4/4 & 23.7 & 22.6 & \textbf{44.7} & 40.1 & 27.2 & 24.4 & 47.7 & 41.1 \\
    RepQ-ViT~\citep{li2023repq} & $\times$ & $\checkmark$ & 4/4 & 36.1 & 36.0 & 44.2 & 40.2 & 47.0 & 41.1 & 49.3 & 43.1 \\
    ERQ~\citep{zhong2024erq} & $\times$ & $\checkmark$ & 4/4 & 36.8 & 36.6 & 43.4 & 40.7 & 47.9 & 42.1 & 50.0 & 43.6 \\
    I\&S-ViT~\citep{zhong2023s} & $\checkmark$ & $\checkmark$ & 4/4 & 37.5 & 36.6 & 43.4 & 40.3 & 48.2 & 42.0 & 50.3 & 43.6 \\
    DopQ-ViT~\citep{yang2024dopq} & $\checkmark$ & $\checkmark$ & 4/4 & 37.5 & 36.5 & 43.5 & 40.4 & 48.2 & 42.1 & 50.3 & 43.7 \\
    QDrop*~\citep{wei2022qdrop} & $\checkmark$ & $\times$ & 4/4 & 36.2 & 35.4 & 41.6 & 39.2 & 47.0 & 41.3 & 49.0 & 42.5 \\
    APHQ-ViT~\citep{wu2025aphq} & $\checkmark$ & $\times$ & 4/4 & 38.9 & 38.1 & 44.1 & 41.0 & \textbf{48.9} & \textbf{42.7} & 50.3 & 43.7 \\
    FIMA-Q~\citep{wu2025fima} & $\checkmark$ & $\times$ & 4/4 & 38.7 & 37.8 & 44.2 & \textbf{41.1} & 48.7 & 42.5 & 50.4 & 43.7 \\
    \textbf{GLF-Q (Ours)} & $\checkmark$ & $\times$ & 4/4 & \textbf{39.6} & \textbf{38.2} & 44.3 & \textbf{41.1} & 48.8 & 42.6 & \textbf{50.6} & \textbf{44.0} \\
    \bottomrule
\end{tabular}%
}
\end{center}
\end{table*}

\subsection{Ablation Studies}
\label{sec:exp_ablation}

\paragraph{On different quantization losses.}
We compare five reconstruction losses within the same 3-bit framework using STE for 3,000 iterations: local feature loss (LF), DPLR-FIM from FIMA-Q~\citep{wu2025fima}, Logits + LF from PD-Quant~\citep{liu2023pd}, global feature loss (GF), and GLF in Eq.~\eqref{eq:glf_loss}. We additionally evaluate FIMA-Q~\citep{wu2025fima} with MLP reconstruction and offline Hadamard transformations using AdaRound for 20,000 iterations, reported separately in Table~\ref{tab:ablation_loss}. As shown in Table~\ref{tab:ablation_loss}, GLF achieves the highest mean accuracy of 72.77\%, exceeding DPLR-FIM, Logits + LF, and GF by 1.38\%, 1.30\%, and 0.98\%, respectively. With MR and Hadamard, FIMA-Q achieves 71.59\% mean accuracy using AdaRound for 20k iterations, 1.17\% below GLF-Q using STE for 3k iterations.

\begin{table*}[t]
\caption{Top-1 accuracy comparison (\%) with different quantization losses and an additional baseline on ImageNet under 3-bit quantization. $\dagger$: FIMA-Q + MR + Hadamard with AdaRound, 20k iterations.}
\label{tab:ablation_loss}
\begin{center}
\begingroup
\setlength{\tabcolsep}{3pt}
\renewcommand{\arraystretch}{1.0}
\begin{tabular*}{\textwidth}{@{\extracolsep{\fill}}l*{7}{c}@{}}
    \toprule
    Setting & ViT-S & ViT-B & DeiT-T & DeiT-S & DeiT-B & Swin-S & Swin-B \\
    \midrule
    LF & 60.48 & 77.06 & 53.16 & 70.16 & 76.76 & 76.66 & 78.46 \\
    DPLR-FIM~\citep{wu2025fima} & 63.83 & 77.15 & 54.94 & 70.67 & 76.60 & 77.30 & 79.19 \\
    Logits + LF~\citep{liu2023pd} & 65.16 & 76.89 & 55.26 & 70.71 & 76.87 & 76.80 & 78.58 \\
    GF & 66.09 & 77.17 & 54.06 & 70.41 & 76.69 & 77.75 & 80.30 \\
    \textbf{GLF (Ours)} & \textbf{67.40} & \textbf{78.18} & \textbf{56.48} & \textbf{71.55} & \textbf{77.20} & \textbf{78.10} & \textbf{80.45} \\
    \midrule
    FIMA-Q + MR + Had.$^{\dagger}$ & 66.88 & 76.13 & 55.29 & 69.81 & 76.24 & 77.37 & 79.42 \\
    \bottomrule
\end{tabular*}
\endgroup
\end{center}
\end{table*}

\paragraph{On main components in GLF-Q.}
To evaluate the contribution of main components in our framework, we conduct an ablation study on MLP Reconstruction and Offline Hadamard Transformation under 3-bit quantization on ImageNet. As shown in Table~\ref{tab:ablation_components}, the full combination achieves the highest accuracy across all evaluated architectures. Adding MR to GLF improves accuracy on five of the seven architectures, with a slight decrease on DeiT-T and no change on DeiT-B. Adding Hadamard transformations improves accuracy across all seven architectures.

\begin{table*}[t]
\caption{Top-1 accuracy comparison (\%) with different framework components on ImageNet under 3-bit quantization. ``GLF'' denotes Global-Local Feature loss, ``MR'' denotes MLP Reconstruction, and ``Hadamard'' denotes Offline Hadamard Transformation.}
\label{tab:ablation_components}
\begin{center}
\begingroup
\setlength{\tabcolsep}{3pt}
\renewcommand{\arraystretch}{1.0}
\begin{tabular*}{\textwidth}{@{\extracolsep{\fill}}l*{7}{c}@{}}
    \toprule
    Method & ViT-S & ViT-B & DeiT-T & DeiT-S & DeiT-B & Swin-S & Swin-B \\
    \midrule
    GLF only & 60.37 & 72.50 & 55.28 & 68.75 & 75.79 & 76.92 & 79.31 \\
    GLF + MR & 63.53 & 73.60 & 55.24 & 69.00 & 75.79 & 77.96 & 80.09 \\
    GLF + Hadamard & 65.04 & 77.16 & 55.69 & 71.30 & 76.91 & 77.63 & 79.81 \\
    \textbf{GLF-Q} & \textbf{67.40} & \textbf{78.18} & \textbf{56.48} & \textbf{71.55} & \textbf{77.20} & \textbf{78.10} & \textbf{80.45} \\
    \bottomrule
\end{tabular*}
\endgroup
\end{center}
\end{table*}

\subsection{Experiments on Out-of-Domain Calibration}
\label{sec:exp_ood}

Table~\ref{tab:ood_sun397} reports ImageNet accuracy after calibration on 1024 unlabeled SUN397 images~\citep{xiao2010sun}. GLF-Q outperforms FIMA-Q~\citep{wu2025fima} on all three models at both bit-widths, with gains of up to 31.97\% on W3A3 DeiT-S. GLF also outperforms DPLR-FIM within our framework at both bit-widths.

\begin{table}[t]
\caption{ImageNet top-1 accuracy (\%) with SUN397 calibration. $\dagger$: DPLR-FIM in our framework.}
\label{tab:ood_sun397}
\begin{center}
\begingroup
\setlength{\tabcolsep}{8.5pt}
\renewcommand{\arraystretch}{1.0}
\begin{tabular}{lcccccc}
    \toprule
    \multirow{2}{*}[-0.5ex]{Method} & \multicolumn{3}{c}{3-bit} & \multicolumn{3}{c}{4-bit} \\
    \cmidrule(lr){2-4} \cmidrule(lr){5-7}
    & ViT-S & DeiT-S & Swin-S & ViT-S & DeiT-S & Swin-S \\
    \midrule
    FIMA-Q~\citep{wu2025fima} & 41.86 & 36.03 & 69.95 & 73.37 & 75.89 & 80.72 \\
    DPLR-FIM$^{\dagger}$ & 47.80 & 65.58 & 70.77 & 73.88 & 76.02 & 80.51 \\
    \textbf{GLF-Q (Ours)} & \textbf{58.72} & \textbf{68.00} & \textbf{74.71} & \textbf{75.34} & \textbf{76.46} & \textbf{81.07} \\
    \bottomrule
\end{tabular}
\endgroup
\end{center}
\end{table}

\subsection{Analysis of Inference Efficiency}
\label{sec:inference_efficiency}

Since quantization below 8 bits requires specialized hardware~\citep{li2021brecq,zhong2024erq}, we evaluate real-world deployment under 8-bit quantization. As shown in Table~\ref{tab:inference_efficiency}, GLF-Q reduces inference latency by 15\% on average compared to FIMA-Q~\citep{wu2025fima}, while improving accuracy across all evaluated models. For ViT and DeiT models, RMSNorm removes the mean-centering operation in LayerNorm, while ReLU simplifies activation computation compared with GELU. Swin models retain LayerNorm but also benefit from the simpler ReLU activation. These computational simplifications can contribute to the observed latency reductions. Deployment and timing details are provided in Appendix~\ref{sec:appendix_implementation}.

\begin{table}[t]
\caption{Comparison of inference latency and Top-1 accuracy under 8-bit quantization. Latency (ms) is measured on a single NVIDIA H800 GPU with a batch size of 64 using NVIDIA TensorRT~\citep{TensorRT}, and Top-1 accuracy is evaluated on ImageNet.}
\label{tab:inference_efficiency}
\begin{center}
\begingroup
\setlength{\tabcolsep}{7pt}
\renewcommand{\arraystretch}{1.0}
\begin{tabular}{lcccccc}
    \toprule
    \multirow{2}{*}[-0.5ex]{Method} & \multicolumn{3}{c}{Latency} & \multicolumn{3}{c}{Top-1 Accuracy} \\
    \cmidrule(lr){2-4} \cmidrule(lr){5-7}
    & ViT-S & DeiT-S & Swin-S & ViT-S & DeiT-S & Swin-S \\
    \midrule
    FIMA-Q~\citep{wu2025fima} & 2.95 & 2.95 & 6.83 & 79.19 & 78.72 & 83.06 \\
    \textbf{GLF-Q (Ours)} & \textbf{2.50} & \textbf{2.49} & \textbf{5.88} & \textbf{80.35} & \textbf{79.07} & \textbf{83.08} \\
    \bottomrule
\end{tabular}
\endgroup
\end{center}
\end{table}

\subsection{Analysis of Training Efficiency}
\label{sec:train_efficiency}

To evaluate practical training efficiency, we compare training time against FIMA-Q~\citep{wu2025fima} and LS-ViT~\citep{hwang2026ls} under 3-bit quantization on ImageNet. All training times are measured on a single NVIDIA H800 GPU. For all methods, the reported time includes calibration and block reconstruction. For GLF-Q, it also includes MLP reconstruction. As shown in Table~\ref{tab:timing_comparison}, GLF-Q is approximately 2.0 to 2.6 times faster than FIMA-Q and 1.6 to 2.2 times faster than LS-ViT across the seven architectures. GLF-Q converges within 3,000 iterations using simple STE, whereas FIMA-Q and LS-ViT use AdaRound with 20,000 reconstruction iterations.

\begin{table}[t]
\caption{Comparison of training time (minutes) on ImageNet under 3-bit quantization.}
\label{tab:timing_comparison}
\begin{center}
\begingroup
\setlength{\tabcolsep}{4pt}
\renewcommand{\arraystretch}{1.0}
\begin{tabular*}{\textwidth}{@{\extracolsep{\fill}}l*{7}{c}@{}}
    \toprule
    Method & ViT-S & ViT-B & DeiT-T & DeiT-S & DeiT-B & Swin-S & Swin-B \\
    \midrule
    FIMA-Q~\citep{wu2025fima} & 48 & 87 & 44 & 49 & 88 & 115 & 135 \\
    LS-ViT~\citep{hwang2026ls} & 40 & 56 & 41 & 40 & 56 & 101 & 104 \\
    \textbf{GLF-Q (Ours)} & \textbf{21} & \textbf{34} & \textbf{19} & \textbf{20} & \textbf{34} & \textbf{57} & \textbf{63} \\
    \bottomrule
\end{tabular*}
\endgroup
\end{center}
\end{table}

\section{Conclusion}
\label{sec:conclusion}

In this paper, we proposed GLF-Q, an efficient post-training quantization framework for Vision Transformers. To address calibration overfitting in logit-guided reconstruction and the approximation limitations of Hessian-based objectives, we introduce the Global-Local Feature loss, which combines penultimate-layer feature alignment through the full-precision downstream network with local reconstruction regularization. Furthermore, offline Hadamard transformations disperse activation outliers to suppress clipping errors with zero runtime overhead, while basic STE optimization achieves rapid convergence. Extensive experiments demonstrate that GLF-Q with standard uniform quantizers substantially outperforms state-of-the-art methods under 3-bit quantization on image classification. GLF-Q also exhibits strong out-of-domain calibration robustness and delivers speedups under 8-bit GPU deployment.

\section*{Acknowledgments and Disclosure of Funding}
This work was partly supported by the National Natural Science Foundation of China under Grant 62276123, by the Fundamental and Interdisciplinary Disciplines Breakthrough Plan of the Ministry of Education of China (No.~JYB2025XDXM118), and by the Zhongguancun Academy Project No. XTS0066.

JW identified problems and proposed conjectures in optimization-based PTQ and guided PS in conducting the experiments. PS, with the help of JW, designed and implemented GLF-Q. GL and JT provided valuable assistance and contributed to discussions on method design and experiments. JW and PS wrote the paper.

\bibliography{iclr2027_conference}
\bibliographystyle{iclr2027_conference}

\clearpage
\appendix
\section*{Appendix}
\section{Uniform Quantizer Formulation}
\label{sec:appendix_uniform_quant}

For efficient hardware deployment, we employ the standard uniform quantizer. Given bit-width $b$, uniform quantization maps a continuous tensor $x$ to its low-bit integer representation $x^{\text{q}}$:
\begin{equation}
x^{\text{q}} = \text{clip} \left( \left\lfloor \frac{x}{s} \right\rceil + z, 0, 2^b - 1 \right)\,,
\end{equation}
where $\lfloor \cdot \rceil$ denotes the round-to-nearest operator, $\text{clip}(\cdot, 0, 2^b - 1)$ restricts values to $[0, 2^b - 1]$, $s \in \mathbb{R}^+$ is the quantization scale, and $z \in \mathbb{Z}$ is the zero-point offset:
\begin{equation}
s = \frac{\max(x) - \min(x)}{2^b - 1}\,, \quad z = \text{clip} \left( \left\lfloor -\frac{\min(x)}{s} \right\rceil, 0, 2^b - 1 \right)\,.
\end{equation}
The dequantized tensor $\hat{x}$ approximates the original tensor $x$ via:
\begin{equation}
\hat{x} = s \times (x^{\text{q}} - z)\,.
\end{equation}

\section{Comprehensive Details and Equivalence Proofs for Offline Hadamard Transformations}
\label{sec:appendix_hadamard}

This section describes the conversion from LayerNorm to RMSNorm, offline Hadamard transformations and their equivalence, and Hadamard construction for dimensions that are not powers of two. We use the row-vector convention throughout.

\subsection{LayerNorm-to-RMSNorm Conversion}
\label{sec:appendix_hadamard_rmsnorm}

Let $C$ denote the channel dimension. Define the mean-centering matrix and centered activation as:
\begin{equation}
M = I - \frac{1}{C}\mathbf{1}\mathbf{1}^{\top} \in \mathbb{R}^{C \times C}\,, \qquad x_{\mathrm{c}} = xM\,,
\end{equation}
where $\mathbf{1} \in \mathbb{R}^C$ is the all-ones vector. LayerNorm can be expressed using affine-free RMSNorm as:
\begin{equation}
\operatorname{LN}(x) = \operatorname{RMSNorm}(xM)\operatorname{Diag}(\gamma) + \beta\,,
\end{equation}
where $\gamma, \beta \in \mathbb{R}^C$ are the affine scaling and bias. RMSNorm retains the original $\epsilon$, while the affine parameters are folded into the subsequent linear layer:
\begin{equation}
\widetilde{W} = \operatorname{Diag}(\gamma)W\,, \qquad \widetilde{b} = \beta W + b\,.
\end{equation}
For a standard pre-LN residual sublayer $x^+ = x + F(\operatorname{LN}(x))$, post-centering yields $x_{\mathrm{c}}^+ = x^+ M = (x + F(\operatorname{LN}(x)))M = x_{\mathrm{c}} + F(\operatorname{LN}(x))M$. To maintain $x_{\mathrm{c},l}=x_lM$ across layers, we fold centering into the patch embedding and the output projections of attention and MLP branches:
\[
\begin{aligned}
E_{\mathrm{patch}} &\leftarrow E_{\mathrm{patch}}M, & b_E &\leftarrow b_EM, \\
W_{\mathrm{out}} &\leftarrow W_{\mathrm{out}}M, & b_{\mathrm{out}} &\leftarrow b_{\mathrm{out}}M.
\end{aligned}
\]
The class token and positional embeddings are also centered as $t\leftarrow tM$ and $p\leftarrow pM$. These updates preserve the features after the final normalization.
\subsection{Residual Hadamard Transformations}
\label{sec:appendix_hadamard_residual}

Following the computational invariance established in SliceGPT~\citep{ashkboos2024slicegpt}, we apply a shared rotation to the residual stream. We select an orthogonal transformation matrix $R_1 \in \mathbb{R}^{C \times C}$ satisfying $R_1^\top R_1 = R_1R_1^\top = I$ and define the rotated residual activation as:
\begin{equation}
x_{\mathrm{c}}^{\mathrm{rot}} = x_{\mathrm{c}}R_1\,.
\end{equation}
Affine-free RMSNorm satisfies orthogonal equivariance~\citep{ashkboos2024quarot}:
\begin{equation}
\operatorname{RMSNorm}(x_{\mathrm{c}}R_1) = \frac{x_{\mathrm{c}}R_1}{\sqrt{\frac{1}{C}\|x_{\mathrm{c}}R_1\|_2^2 + \epsilon}} = \frac{x_{\mathrm{c}}R_1}{\sqrt{\frac{1}{C}\|x_{\mathrm{c}}\|_2^2 + \epsilon}} = \operatorname{RMSNorm}(x_{\mathrm{c}})R_1\,,
\end{equation}
because orthogonal transformations preserve Euclidean norms. Table~\ref{tab:hadamard_rules} summarizes the corresponding offline parameter transformations.

\begin{table}[h]
\caption{Parameter transformation rules for the centered and affine-folded Vision Transformer.}
\label{tab:hadamard_rules}
\begin{center}
\begingroup
\small
\setlength{\tabcolsep}{4pt}
\renewcommand{\arraystretch}{1.1}
\begin{tabular*}{\linewidth}{@{\extracolsep{\fill}}ll@{}}
\toprule
Network Position & Parameter Transformation Rule \\
\midrule
Input Patch Embedding & $E_{\mathrm{patch}} \to E_{\mathrm{patch}}R_1\,, \quad b_E \to b_E R_1$ \\
Class Token \& Positional Embeddings & $t \to tR_1\,, \quad p \to pR_1$ \\
QKV Projection \& MLP First Layer & $W_{\mathrm{in}} \to R_1^{\top}W_{\mathrm{in}}\,, \quad b_{\mathrm{in}} \text{ (unchanged)}$ \\
Attention Output Projection \& MLP Second Layer & $W_{\mathrm{out}} \to W_{\mathrm{out}}R_1\,, \quad b_{\mathrm{out}} \to b_{\mathrm{out}}R_1$ \\
Final Classification Head & $W_{\mathrm{head}} \to R_1^{\top}W_{\mathrm{head}}\,, \quad b_{\mathrm{head}} \text{ (unchanged)}$ \\
\bottomrule
\end{tabular*}
\endgroup
\end{center}
\end{table}

Input and output projections absorb $R_1^\top$ and $R_1$, respectively, so both residual branches share the same rotation. RMSNorm equivariance allows this rotation to propagate through the network, while the classification head absorbs its inverse, preserving the logits.
\subsection{Head-wise QK and VO Transformations}
\label{sec:appendix_hadamard_locations}
\paragraph{Head-wise QK Rotation.}
Let $Q_h,K_h\in\mathbb{R}^{N\times C_h}$ denote the query and key states of attention head $h$, where $C_h=C/N_h$. For an orthogonal rotation $R_{2,h}\in\mathbb{R}^{C_h\times C_h}$, we define $Q'_h=Q_hR_{2,h}$ and $K'_h=K_hR_{2,h}$, with the rotation folded into the query and key projections offline. Then
\begin{equation}
Q'_h {K'_h}^{\top} = (Q_h R_{2,h})(K_h R_{2,h})^{\top} = Q_h R_{2,h} R_{2,h}^{\top} K_h^{\top} = Q_h K_h^{\top}\,.
\end{equation}
Because the inner products entering Softmax remain unchanged, the resulting attention probabilities are also unchanged.

\paragraph{Head-wise VO Rotation.}
Let $P_h \in \mathbb{R}^{N \times N}$ denote the attention probability matrix, $V_h \in \mathbb{R}^{N \times C_h}$ the value states, and $W_{O,h} \in \mathbb{R}^{C_h \times C}$ the row slice of the output projection corresponding to head $h$. Under the row-vector convention, this head contributes $(P_h V_h)W_{O,h}$ to the attention output. For an orthogonal rotation $R_{3,h} \in \mathbb{R}^{C_h \times C_h}$, we define $V'_h = V_h R_{3,h}$ and fold the inverse rotation into the output projection as $W'_{O,h} = R_{3,h}^\top W_{O,h}$. Then
\begin{equation}
(P_h V'_h)W'_{O,h} = (P_h V_h R_{3,h})(R_{3,h}^\top W_{O,h}) = P_h V_h W_{O,h}\,.
\end{equation}
Summing these contributions over all heads and retaining the output projection bias preserves the multi-head self-attention output.

\subsection{Randomized Hadamard Construction}
\label{sec:appendix_hadamard_construction}

Following QuaRot~\citep{ashkboos2024quarot}, we construct normalized randomized Hadamard matrices for the dimensions used in our models. For $d=2^k$, we start with $H_1=[1]$ and recursively construct
\[
H_{2n}=\begin{bmatrix}
H_n & H_n \\
H_n & -H_n
\end{bmatrix}
\]
until the dimension reaches $d$. We then define the normalized randomized Hadamard matrix as
\begin{equation}
R_d = \frac{DH_d}{\sqrt{d}}\,,
\end{equation}
where $D$ is a diagonal matrix with independent entries uniformly sampled from $\{-1,+1\}$.

For $d=12\cdot 2^k$, we combine a fixed Hadamard matrix $H_{12}$ with $H_{2^k}$ using the Kronecker product:
\begin{equation}
R_d = \frac{D\left(H_{12}^{\top}\otimes H_{2^k}\right)}{\sqrt{d}}\,.
\end{equation}
This construction supports hidden dimensions such as 192, 384, and 768 without padding or truncation. Since $H_m^{\top}H_m=mI_m$ and $D^{\top}D=I_d$, both constructions satisfy $R_d^{\top}R_d=I_d$.
\section{Overall Pipeline of GLF-Q}
\label{sec:appendix_pipeline}

The overall pipeline of GLF-Q is summarized in Algorithm~\ref{alg:glfq_pipeline}.

\begin{algorithm}[H]
\caption{Overall Pipeline of GLF-Q}
\label{alg:glfq_pipeline}
\begin{algorithmic}[1]
\REQUIRE Pretrained model $\mathcal{M}$; unlabeled calibration data $\mathcal{D}_{\mathrm{cal}}$; bit-widths $(b_w,b_a)$; reconstruction iterations $T=3000$; regularization coefficient $\lambda=2$.
\ENSURE Quantized model $\widehat{\mathcal{M}}$.
\STATE Initialize teacher $\mathcal{M}_{\mathrm{fp}}$ and student $\mathcal{M}_{\mathrm{s}}$ from $\mathcal{M}$.
\STATE Apply identical offline transformations to both models following Section~\ref{sec:pipeline} and the architecture-specific settings in Section~\ref{sec:exp_setup}.
\STATE Freeze $\mathcal{M}_{\mathrm{fp}}$; replace student MLP GELU with ReLU and reconstruct against $\mathcal{M}_{\mathrm{fp}}$.
\STATE Insert and calibrate student uniform quantizers; apply the Swin reparameterization when applicable.
\FOR{each reconstruction block $l$ in forward order}
    \STATE Cache student inputs $\hat{x}_l$ from the quantized prefix and teacher targets $(y_l,f)$ on $\mathcal{D}_{\mathrm{cal}}$.
    \STATE Let $F_l$ be the frozen teacher suffix; optimize only the current student block.
    \FOR{$t=1,\ldots,T$}
        \STATE Sample matching batches of $(\hat{x}_l,y_l,f)$.
        \STATE Compute the current quantized block output $\hat{y}_l$ from $\hat{x}_l$ and obtain $\hat{f}=F_l(\hat{y}_l)$.
        \STATE Evaluate $\mathcal{L}_{\mathrm{global}}^{(l)}$ and $\mathcal{L}_{\mathrm{local}}^{(l)}$ on the minibatch using Eqs.~\eqref{eq:global_feature_loss} and~\eqref{eq:local_feature_loss}.
        \IF{$t=1$}
            \STATE Cache the detached loss values, each clamped below at $10^{-12}$, as fixed normalization factors.
        \ENDIF
        \STATE Normalize both loss terms by their cached factors and combine them using Eq.~\eqref{eq:glf_loss}.
        \STATE Backpropagate through $F_l$ and the quantizers using STE; update $\theta_l$ and $s_l$ with Adam.
    \ENDFOR
\ENDFOR
\end{algorithmic}
\end{algorithm}

\section{Effect of Loss Functions on Generalization}
\label{sec:appendix_overfitting}

To evaluate generalization under limited calibration data, Table~\ref{tab:calibration_validation_gap} compares the top-1 accuracy of logit and feature supervision on the calibration set and the ImageNet validation set. Logits + LF combines logit supervision with local feature regularization from PD-Quant~\citep{liu2023pd}, while GLF denotes the proposed Global-Local Feature loss. Both variants follow the 3-bit setting in Table~\ref{tab:ablation_loss} and differ only in the loss function.

\begin{table}[htbp]
\caption{Top-1 accuracy (\%) on the calibration set and ImageNet validation set under 3-bit quantization.}
\label{tab:calibration_validation_gap}
\begin{center}
\begingroup
\setlength{\tabcolsep}{8pt}
\renewcommand{\arraystretch}{1.05}
\begin{tabular}{lcccc}
\toprule
& \multicolumn{2}{c}{Logits + LF} & \multicolumn{2}{c}{GLF} \\
\cmidrule(lr){2-3}\cmidrule(lr){4-5}
Model & Calibration & Validation & Calibration & Validation \\
\midrule
ViT-S  & 86.91 & 65.16 & 87.70 & 67.40 \\
ViT-B  & 91.11 & 76.89 & 91.11 & 78.18 \\
DeiT-T & 77.05 & 55.26 & 76.66 & 56.48 \\
DeiT-S & 90.04 & 70.71 & 89.65 & 71.55 \\
DeiT-B & 96.00 & 76.87 & 96.00 & 77.20 \\
Swin-S & 87.40 & 76.80 & 87.70 & 78.10 \\
Swin-B & 92.38 & 78.58 & 92.48 & 80.45 \\
\midrule
Mean   & 88.70 & 71.47 & 88.76 & 72.77 \\
\bottomrule
\end{tabular}
\endgroup
\end{center}
\end{table}

With nearly identical mean calibration accuracy (88.76\% for GLF and 88.70\% for Logits + LF), GLF yields higher validation accuracy across all seven architectures. On average, GLF improves validation accuracy by 1.30\% and narrows the performance gap between the calibration and validation sets from 17.23\% to 15.99\%. These results suggest better generalization with GLF under limited calibration data.

\section{Stability across Random Seeds}
\label{sec:appendix_multiseed}

Table~\ref{tab:multiseed} reports multi-seed ImageNet top-1 accuracy for GLF-Q across seven architectures under 3-bit and 4-bit quantization.

\begin{table}[htbp]
\caption{Multi-seed top-1 accuracy (\%) of GLF-Q on ImageNet under 3-bit and 4-bit quantization, reported as mean $\pm$ standard deviation over three random seeds.}
\label{tab:multiseed}
\begin{center}
\resizebox{\textwidth}{!}{%
\begin{tabular}{cccccccc}
\toprule
\textbf{W/A} & \textbf{ViT-S} & \textbf{ViT-B} & \textbf{DeiT-T} & \textbf{DeiT-S} & \textbf{DeiT-B} & \textbf{Swin-S} & \textbf{Swin-B} \\
\midrule
3/3 & $67.34 \pm 0.12$ & $78.21 \pm 0.04$ & $56.17 \pm 0.27$ & $71.67 \pm 0.10$ & $77.22 \pm 0.10$ & $78.07 \pm 0.06$ & $80.44 \pm 0.03$ \\
4/4 & $77.69 \pm 0.15$ & $83.12 \pm 0.10$ & $67.22 \pm 0.16$ & $77.40 \pm 0.06$ & $80.51 \pm 0.04$ & $82.05 \pm 0.08$ & $83.81 \pm 0.01$ \\
\bottomrule
\end{tabular}
}
\end{center}
\end{table}

\section{Additional Ablation Studies}
\label{sec:appendix_additional_ablations}

\paragraph{Sensitivity to the regularization coefficient.}
We evaluate GLF-Q with $\lambda\in\{1,2,4\}$ under 3-bit quantization on ImageNet. As shown in Table~\ref{tab:lambda_sensitivity}, mean top-1 accuracy across seven architectures ranges from 72.77\% to 72.81\%. These results suggest limited sensitivity to $\lambda$ within the evaluated range.

\begin{table}[htbp]
\caption{Top-1 accuracy (\%) of GLF-Q with different regularization coefficients under 3-bit quantization on ImageNet.}
\label{tab:lambda_sensitivity}
\begin{center}
\begingroup
\small
\setlength{\tabcolsep}{4pt}
\renewcommand{\arraystretch}{1.05}
\begin{tabular*}{\textwidth}{@{\extracolsep{\fill}}c*{8}{c}@{}}
\toprule
$\lambda$ & ViT-S & ViT-B & DeiT-T & DeiT-S & DeiT-B & Swin-S & Swin-B & Mean \\
\midrule
1 & 67.53 & 78.29 & 56.31 & 71.47 & 77.33 & 78.22 & 80.49 & 72.81 \\
2 & 67.40 & 78.18 & 56.48 & 71.55 & 77.20 & 78.10 & 80.45 & 72.77 \\
4 & 67.48 & 78.18 & 56.61 & 71.59 & 77.22 & 78.14 & 80.35 & 72.80 \\
\bottomrule
\end{tabular*}
\endgroup
\end{center}
\end{table}

\clearpage
\section{Implementation Details}
\label{sec:appendix_implementation}

\subsection{MLP Reconstruction}
\label{sec:appendix_mlp_reconstruction}

MLP reconstruction is performed block-wise in floating-point precision before quantization to compensate for the error introduced by replacing GELU with ReLU. The teacher retains GELU, while the student uses ReLU. Both use identical coordinate transformations. Each student MLP is optimized using cached teacher inputs $x$ and outputs $y$ with the objective
\begin{equation}
\mathcal{L}_{\mathrm{MR}}
= \frac{1}{NC}\left\|f_{\mathrm{ReLU}}(x)-y\right\|_F^2
+ \frac{2}{NC}\left\|f_{\mathrm{ReLU,clip}}(x)-y\right\|_F^2,
\label{eq:mlp_reconstruction_loss}
\end{equation}
where $N$ and $C$ denote the number of output tokens and channels, respectively. The function $f_{\mathrm{ReLU,clip}}$ applies upper clipping to the student's FC2 input, with a fixed threshold for each block set to the 99th percentile of positive activations at the corresponding teacher FC2 input over the calibration data.

Each MLP is optimized for 20,000 iterations using Adam without weight decay. The learning rate follows an independent cosine schedule for each block, decaying from $4\times10^{-5}$ to zero. Only the weights and biases of the current student MLP's FC1 and FC2 layers are updated. All other parameters remain frozen.

\subsection{Object Detection and Segmentation}
\label{sec:det_loss}

To extend GLF-Q to object detection and instance segmentation, we use intermediate bbox and mask RoI features before the prediction layers as global supervision targets. The bbox and mask branches use up to 1000 and 128 student RPN proposals per image, respectively, preserving their original order. Teacher and student paths use identical RoI coordinates for each loss. The proposals and teacher reference features are cached once and reused throughout reconstruction.

Backbone reconstruction combines local, bbox RoI, and mask RoI feature MSE losses with weights of 2, 0.5, and 0.5, respectively. Each loss term is normalized by its value at the first reconstruction iteration, with the normalization factor held fixed for that block. FPN reconstruction uses only the normalized local MSE with a weight of 2, while the RPN and RoI heads undergo calibration-only quantization.

\subsection{TensorRT Deployment and Latency Measurement}
\label{sec:appendix_tensorrt}

To support TensorRT's standard INT8 path, asymmetric quantization representations that are not directly compatible are exported in floating-point form and requantized using a common scheme. We use symmetric absmax quantization per output channel for weights and per tensor for activations.

All latencies are measured on the same NVIDIA H800 80GB GPU using TensorRT 10.9, with an input resolution of $224\times224$ and a batch size of 64. After 50 warm-up runs, we time 200 inference runs using CUDA events and report the mean batch latency. Timing covers only GPU engine execution, excluding data loading, preprocessing, and data transfers. Engines are built with INT8 enabled and FP32/TF32 fallback allowed.

\end{document}